\documentclass[letterpaper, 10 pt, conference]{ieeeconf}  

\IEEEoverridecommandlockouts                              

\usepackage{graphics} 
\usepackage{epsfig} 
\usepackage{amsmath} 
\usepackage{subfigure}
\usepackage{multirow}
\usepackage{array}

\title{\LARGE \bf
Localizing Discriminative Visual Landmarks for Place Recognition
}

\author{Zhe Xin$^{12}$, Yinghao Cai$^{1}$, Tao Lu$^{1}$, Xiaoxia Xing$^{12}$, Shaojun Cai$^{3}$,  \\ Jixiang Zhang$^{1}$, Yiping Yang$^{1}$, Yanqing Wang$^{1}$     
\thanks{$^{1}$Institute of Automation, Chinese Academy of Sciences, Beijing, China 
{\tt\small \{yinghao.cai,tao.lu,jixiang.zhang,
yiping.yang,yanqing.wang\}@ia.ac.cn}}
\thanks{$^{2}$University of Chinese Academy of Sciences, Beijing, China
{\tt\small \{xinzhe2015,xingxiaoxia2015\}@ia.ac.cn}}
\thanks{$^{3}$UISEE Technologies Beijing Co., Ltd
{\tt\small \{shaojun.cai\}@uisee.com}}
}

\begin{document}

\maketitle
\thispagestyle{empty}
\pagestyle{empty}

\begin{abstract}

We address the problem of visual place recognition with perceptual changes. The fundamental problem of visual place recognition is generating robust image representations which are not only insensitive to environmental changes but also distinguishable to different places. Taking advantage of the feature extraction ability of Convolutional Neural Networks (CNNs), we further investigate how to localize discriminative visual landmarks that positively contribute to the similarity measurement, such as buildings and vegetations. In particular, a Landmark Localization Network (LLN) is designed to indicate which regions of an image are used for discrimination. Detailed experiments are conducted on open source datasets with varied appearance and viewpoint changes. The proposed approach achieves superior performance against state-of-the-art methods. 

\end{abstract}

\section{INTRODUCTION}

Visual place recognition aims to localize the query image by finding the most similar images stored in a pre-built environmental map. Since vision is the primary sensor for many robotic applications, visual place recognition has made great progress in recent years. However, visual place recognition is still an open problem. In long-term robot autonomy, varied illumination and weather conditions, viewpoint changes and dynamic objects lead the same place appear dramatically different. Instead of perceptual changes, the difficulty of place recognition also comes from confusing visual elements, such as sky and roads. These misleading elements make different places indistinguishable. All these variations increase the difficulties of visual place recognition and make it still one of the most challenging tasks in robotic applications.

The key component of visual place recognition is how to describe a specific place against various perceptual changes~\cite{survey}. Generally speaking, image representation learning can fall into two main categories in visual place recognition. The first category describes the whole image using a holistic feature~\cite{conv3,NetVLAD,segment,oxfordtrain,scale}. The second one describes the image with a set of local features. Although holistic features are much more robust against appearance changes, they are sensitive to viewpoint changes and partial occlusions~\cite{survey}. On the contrary, local features are much more viewpoint invariant and show the capability to deal with partial occlusions. Recently, highly representative local Convolutional Neural Networks (CNNs) features~\cite{landmarks,once,superpixel} have been demonstrated to outperform traditional Bag-of-Words (BoWs) models~\cite{dbow,fabmap} on visual place recognition.

\begin{figure}[t]
\centering
\begin{tabular}{c}
\includegraphics[width=7cm]{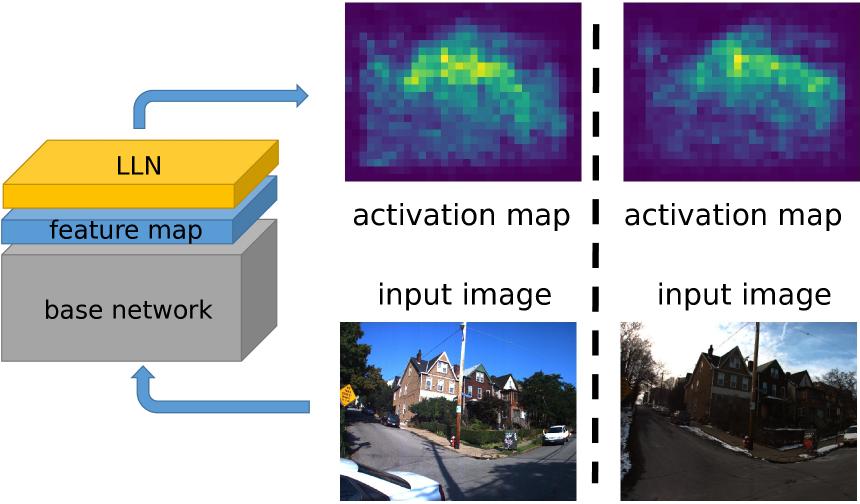}
\end{tabular}
\caption{Landmark localization network is designed to localize discriminative visual landmarks for place recognition.}
\label{fig:f1}
\vspace{-5.5mm}
\end{figure}

\begin{figure*}[t]
\centering
\begin{tabular}{c}
\includegraphics[width=15cm]{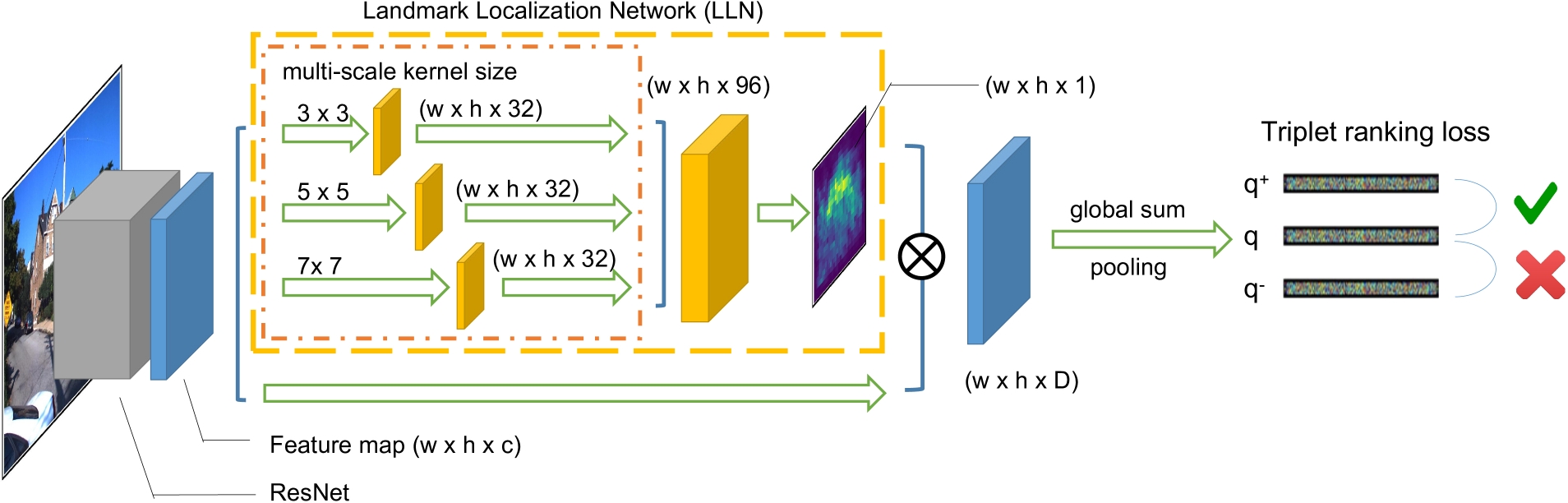}
\end{tabular}
\caption{The overall illustration of the proposed network.}
\label{fig:f2}
\vspace{-5.5mm}
\end{figure*}

However, it is noted that not all local regions are playing equally important in visual place recognition. For example, buildings are quite discriminative and stable against various perceptual changes when used for describing places in visual place recognition. While road surfaces are hardly distinguishable to different places and may bring ambiguities in similarity measurement. Therefore, it is important to identify discriminative image regions to define a particular place from which local features can be extracted. Sunderhauf et al.~\cite{landmarks} utilized edges~\cite{edgeboxes} to discover object proposals. Each object proposal is described with CNN-based features. However, external detectors like edge boxes~\cite{edgeboxes} rely on low-level vision structure without considering the context information of images. Chen et al.~\cite{once} extracted visual landmarks directly based on activations of the convolutional layers. The utilized network was pre-trained for the task of object recognition. The extracted landmarks in~\cite{once} may not be representative for identitying specific places.


Kim et al.~\cite{reweight} learned to generate image representations incorporating context-aware feature preponderance. Naseer et al.~\cite{segment} employed semantic segmentation to extract meaningful features from buildings. However, the method in~\cite{segment} relies on supervised priors, which is of limited use if the scenes do not contain categories in semantic segmentation. All the above approaches~\cite{reweight,segment} generate holistic features to describe the entire image, which lack of the ability for dealing with partial occlusion and background clutters. Recently, Noh et al.~\cite{delf} proposed an attention based network named DELF for the task of image retrieval. Coupled with the attention mechanism, DELF could generate semantic local features. They interpret the task as an classification problem and train the network with a cross-entropy loss. However, place recognition is different from image classification, since the former emphasizes on recognizing the most similar images of a unique place rather than images belonging to the same category.



In this paper, we propose a novel convolutional neural network to identify discriminative image regions for place recognition. Metric learning scheme and hard negative mining strategy are employed to train the proposed network. The metric loss can effectively measure the similarity between places in various environmental conditions. We describe each image with a set of discriminative landmarks to take advantage of local invariant features.

More specifically, a Landmark Localization Network (LLN) is designed to generate an activation map which indicates the saliency of each local feature in the feature map, as shown in Fig.~\ref{fig:f1}. The LLN is trained in an end-to-end manner with only image-level annotations. Each local feature, which is regarded as a visual landmark, corresponds to a local region in the image. The higher the activation of a local feature is, the more representative the landmark is to describe the image. The similarity between two images is obtained by crossly matching of these discriminative landmarks.


Instead of labeling stable or non-stable regions in the training images, discriminative regions are discovered in a weakly supervised way with only image-level annotations. The network heuristically learns to localize landmarks that positively contribute to identifying unique places. The selected landmarks are not only restricted to different perspectives of buildings but also include vegetations and man-made structures.

We evaluate the proposed method on several datasets with variations in both viewpoint and appearance. Our proposed method achieves superior performance against other state-of-the-art methods. Moreover, we integrate the whole process into ORB-SLAM~\cite{orbslam} to verify its performance on autonomous driving applications. Extensive experimental results demonstrate that the proposed approach can effectively improve the localization performance with monthly even seasonal changes.

\section{Methodology}
In this section, the proposed approach is described in detail. We first illustrate the structure of our Landmark Localization Network (LLN) in Section~\ref{lln}. Then, the training process is introduced in Section~\ref{training}. Finally, we present the similarity measurement process in Section~\ref{sim}. The overall structure of the proposed network is shown in Fig.~\ref{fig:f2}.    


\subsection{Landmark Localization Network}
\label{lln}
The landmark localization network is integrated with a base network for dense feature extraction. Local features represent image regions based on their receptive fields. For example, the feature map with a dimensional of $w\times h\times C$ is treated as a set of $C$ dimensional local features extracted at $w\times h$ locations. We employ the ResNet101~\cite{resnet} model as our base network. The output feature map of the third convolutional block is used as the input to LLN.

The output of the LLN is an activation map which indicates the discriminativeness of each local feature for describing the image. This is implemented by using several convolutional filters, as shown in Fig.~\ref{fig:f2}. In order to generate better invariance against viewpoint changes and partial occlusions, we employ multi-scale filters with different kernel sizes. The outputs of multi-scale filters are concatenated to a feature map with $w\times h\times D$. $D$ equals the sum of all multi-scale filters. Then, the activation map is generated by combining all activations of each spatial location together with $1\times 1\times D$ convolutional layer followed by ReLU activation. The ReLU operation ensures all values of the activation map to be non-negative. The activation map localizes discriminative image regions used for image description.


\subsection{Training}
\label{training}
The network is trained with only image-level annotations. In order to train the LLN in an end-to-end manner. An embedding of the whole image needs to be generated for calculating the metric loss. This is done by a weighted sum of all local features $f_{ij}$, which is given by
\begin{equation} 
F = \sum_{i}\sum_{j}w_{ij}\times f_{ij}
\end{equation}
where $i$ = $1, ..., w$ and $j$ = $1, ..., h$. The corresponding weight $w_{ij}$ is predicted by LLN. After aggregating, the feature is L2 normalized for metric learning.

We use the Retrieval-SFM dataset~\cite{trainingdataset}, which is widely used for instance image retrieval. The dataset contains $28559$ images from $713$ locations in the world. Each image has a label indicating the location it belongs to. Most locations are famous man-made architectures such as palaces and towers, which are relatively static and positively contribute to visual place recognition. The training dataset contains various perceptual changes including variations in viewing angles, occlusions and illumination conditions, etc.

We set images belonging to the same location as matched pairs while images from different locations are regarded as non-matched pairs. The objective of learning the LLN is to ensure the distance of feature representations between matched pairs is smaller than non-matched pairs. Since the perceptual changes between matched pairs are severe, LLN can learn to localize the most discriminative local features to represent images. 

We choose $5974$ matched pairs in the Retrieval-SFM dataset for training. All positive images are chosen based on relaxed inliers as described in~\cite{trainingdataset}. The non-matched pairs are generated by hard negative mining. For each query image, we rank all dataset images by comparing the aggregated features. Then, the non-matched images are the top $K$ images which belong to different locations but similar to the query image. In the training process, non-matched images are selected at the beginning of each epoch. The top number $K$ is set to $4$ in our experiments.

\begin{figure}[t]
\centering
\begin{tabular}{c}
\includegraphics[width=8.0cm]{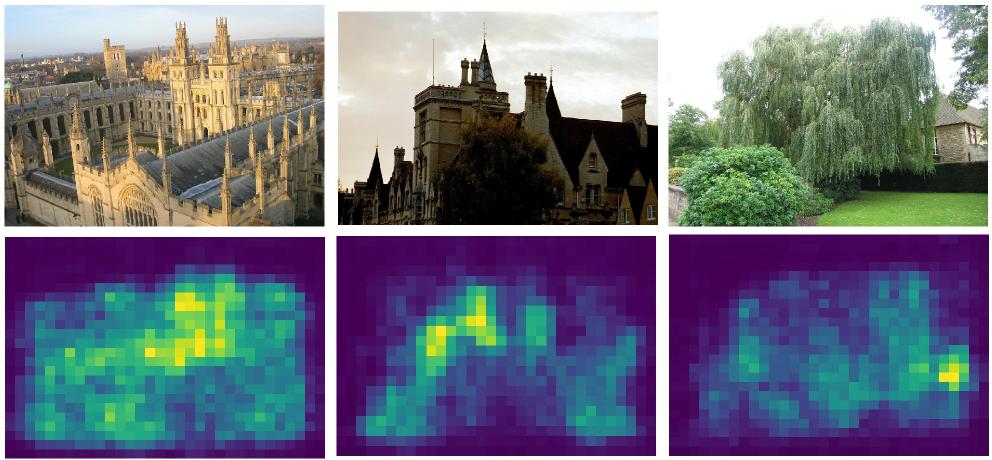}
\end{tabular}
\caption{Some training images and their activation maps generated by the proposed network. For best visualization, images are fed into the network with their original resolution.}
\label{fig:f3}
\vspace{-5.5mm}
\end{figure}

We provide image tuples to the network, each tuple contains $6$ images, including $1$ query image $q$, $1$ matched image $q^+$ and $4$ non-matched images $q^-$. The loss function is based on triplet ranking loss:

\vspace{-3mm}
\begin{equation}
L = \sum_i^K max(0, ||F_q - F_{q^+}||_2 + m - ||F_q - F_{q_i^-}||_2).
\end{equation}
where $K$ is the number of the non-matched image pairs. $m$ stands for margin. By minimizing the loss function in (2), the network can learn to allocate higher activations for discriminative visual landmarks. The whole process is heuristic that no pairwise matching/non-matching annotations are necessary. Some training images and their activation maps are shown in Fig.~\ref{fig:f3}. It can be observed in Fig.~\ref{fig:f3} that LLN can effectively localize discriminative landmarks to describe the image.

\subsection{Similarity Measurement}
\label{sim}
Based on the result of LLN, we select top $n$ local features with the highest activations to generate image representations. Each local feature describes a region of the input image. The size of each region relies on the receptive field of the local feature. The similarity between two images $A$ and $B$ is determined by a weighted sum of their matched local features. More specifically, let $I_1$, $I_2$ be two images with landmarks {$l_1^i, l_2^i, ..., l_n^i$}, where $i$ = $1,2$. We perform cross matching to obtain mutually matched landmark pairs. The similarity between $l_i^1$ and $l_j^2$ is calculated as follows:

\begin{equation}
s_{ij} = \frac{f(l_i^1)^Tf(l_j^2)}{||f(l_i^1)||||f(l_j^2)||}.
\end{equation}

To eliminate the influence of outliers, the image similarity is the weighted sum of all mutually matched landmark pairs. The weight is calculated based on the spatial distribution of each match. Here, we use the center of each region as its coordinate. Generally, correct matches should have similar coordinate differences and incorrect matches have random coordinate differences~\cite{match,gms}. Therefore, a two-dimensional histogram is built by analyzing the coordinate differences in both $x$ and $y$ direction of all landmark pairs. Then, we select the difference with the highest frequency, denoted as ($d_x$, $d_y$). Let ($x_i^1$, $y_i^1$) and ($x_j^2$, $y_j^2$) be the coordinates of two landmarks $l_i^1$ and $l_j^2$, the weight is given by:

\vspace{-3mm}
\begin{equation}
ws_{ij} =  exp(-\frac{1}{2}(((x_i^1 - x_j^2) - d_x)^2 + ((y_i^1 - y_j^2) - d_y)^2)).
\end{equation}
\vspace{-3mm}

The larger the difference compared to the statistics, the smaller the weight is. It should be noted that we make the assumption that views of places are captured from the same direction, which is practical in autonomous driving applications. Overall, the image similarity is calculated as follows:

\vspace{-3mm}
\begin{equation}
S_{12} = \sum_{i,j}ws_{ij} \times s_{ij}.
\end{equation}
\vspace{-3mm}

\section{Experimental Setup}
Our experiments are composed of two parts. First, we discuss the performance of the proposed approach compared to state-of-the-art descriptors. Second, we integrate the whole process in a localization system based on ORB-SLAM~\cite{orbslam} to verify the improvement of relocalization performance in severe environmental changes.

\subsection{Implementation Details}
The base network is ResNet101 pre-trained on ImageNet, the feature maps are extracted from the third convolutional block. LLN is initialized using Xavier initializer. In the training process, we only train LLN and keep the base network unchanged. The reason is that Retrieval-SFM dataset~\cite{trainingdataset} has relatively fewer appearance changes than viewpoint variations. Fine-tuning the base network may lose its robustness to severe environmental changes. The training of LLN is more about learning how to emphasize on place-specific and stable visual landmarks. On the contrary, CNN features trained on ImageNet dataset have been demonstrated to achieve satisfactory results in various environmental changes. 


The proposed network is fully convolutional and not restricted to the image size. However, we resize all training images to $352\times 352$ for training efficiency. In testing, all images are resized to have a height of $288$, while keeping the aspect ratio unchanged. We use Adam solver~\cite{adam} and set the initial learning rate to $1\times 10^{-6}$. The margin is $0.3$. The batch size is $10$. Our implementation is based on Tensorflow.

\subsection{Place Recognition}
\subsubsection{\bf{Comparison and Evaluation Protocol}}
The proposed approach is compared with several state-of-the-art holistic and local descriptors.

\begin{itemize}

\item {\bf{Holistic}}. This descriptor is the aggregation of the base network feature map. We use global max pooling to generate a $1024$ dimensional holistic feature.

\item {\bf{LLN}}. Top $K$ landmarks are selected based on LLN. Each local feature has a dimension of 1024. Only the most discriminative visual elements are used to represent the image.

\item {\bf{ACT}}. Similar to Chen et al.~\cite{once}, we also extract top $K$ landmarks based on activations of the base network. The statistics of the feature map can reflect which part of the image that the network focuses on.

\item {\bf{RAND}}. Top $K$ landmarks are randomly selected without using LLN. We use the same random locations for both query and map images, regardless of viewpoint changes of the testing datasets.

\item {\bf{ALL}}. Without selection, densely sampled local features are used for the similarity measurement. 

\item {\bf{DELF~\cite{delf}}}. The original input of DELF is a set of multi-scale images, we only use the original image scale in our experiments. All other settings are followed the same as~\cite{delf}. We use the source code publicly available.

\item {\bf{Conv3~\cite{conv3}}}. This descriptor is the activation of the third convolutional layer of AlexNet~\cite{alexnet}. Conv3 is a commonly used holistic feature in visual place recognition. 

\end{itemize}

Given a query image, we search the best matching reference image by ranking all images in the map and pick the one with the highest similarity score. Since measuring image similarity based on local features is relatively time-consuming than holistic features, we eliminate many unsimilar images to make the measurement of local features much more efficient. Instead of ranking all map images, \textbf{LLN}, \textbf{ACT}, \textbf{RAND}, \textbf{ALL} and \textbf{DELF} only rerank the top 30 results of the \textbf{Holistic} method. Although the holistic descriptor is sensitive to viewpoint changes, it is robust enough to exclude wrong places, as shown in Fig.~\ref{fig:f4}.

We use Precision-Recall curve to benchmark visual place recognition. The result is regarded as correct if the top $1$ map image is within a vision offset of the reference image. The generation of Precision-Recall curve is adopted from~\cite{conv3}.

\begin{table*}[t]
\caption{Summarization of testing datasets.}
\vspace{-4mm}
\label{tab:t1}
\begin{center}
\begin{tabular}{|cccccccccc|}
\hline 
 & & &\multicolumn{2}{c}{\bf{Variations in}}& &\multicolumn{2}{c}{\bf{Resolution}} & \multicolumn{2}{c|}{\bf{Landmarks}} \\
\multirow{2}{*}{\bf{Datasets}} & \multirow{2}{*}{\bf{Sequences}} & \bf{No. of} & \multirow{2}{*}{\bf{Appearance}} & \multirow{2}{*}{\bf{Viewpoint}} & \bf{Vision} & \multirow{2}{*}{\bf{Original}} & \multirow{2}{*}{\bf{Resized}} & \multirow{2}{*}{\bf{Selected}} & \multirow{2}{*}{\bf{Total}} \\
 & & \bf{frames} & & & \bf{offset} & & & & \\ 
\hline
\multirow{2}{*}{Gardens Point} & day\_left vs. night\_right & \multirow{2}{*}{400} & Severe & Severe & \multirow{2}{*}{3} & \multirow{2}{*}{960$\times$540} & \multirow{2}{*}{512$\times$288} & \multirow{2}{*}{75} & \multirow{2}{*}{144} \\
& day\_right vs. night\_right & & Moderate & Severe & & & & & \\
\hline
\multirow{3}{*}{CMU} & summer vs. fall & \multirow{3}{*}{224} & Moderate & Moderate & \multirow{3}{*}{1} & \multirow{3}{*}{1024$\times$768} & \multirow{3}{*}{384$\times$288} & \multirow{3}{*}{50} & \multirow{3}{*}{108}\\
& summer vs. winter & & Severe & Moderate & & &  & & \\
& fall vs. winter & & Severe & Moderate & & &  & & \\
\hline
Mapillary & cloudy vs. sunny & 366 & Moderate & Moderate & 2 & 640$\times$480 & 384$\times$288 & 50 & 108\\
\hline
\end{tabular}
\end{center}
\vspace{-3mm}
\end{table*}

\subsubsection{\bf{Datasets}}

The proposed approach is evaluated on three place recognition datasets, Gardens Point Walking dataset\footnote{https://wiki.qut.edu.au/pages/viewpage.action?pageId=175739622}, CMU Localization dataset\footnote{http://3dvis.ri.cmu.edu/data-sets/localization} and Mapillary dataset\footnote{https://www.mapillary.com/}. These datasets contain several traverses and capture various appearance and viewpoint changes. Gardens Point Walking dataset is recorded on Gardens Point Campus of QUT. There are three traverses along the same route, two during the day and one during the night. The two day traverses are captured on the left and right side of the pathway, respectively (day\_left, day\_right). The night traverse is captured on the right side of the pathway (night\_right). CMU Localization dataset consists of traverses along the same route around Pittsburgh (USA) during different seasons and years. We use images captured from the left mono camera of the car and select three traverses 01/09/2010 (summer), 28/10/2010 (fall) and 21/12/2010 (winter) to carry out experiments. Mapillary is a crowdsourced photo-mapping platform. We select two traverses captured by dashboard cameras on cars in Malmo, Sweden. Both traverses are recorded from the same lane on a road, one from a sunny day (sunny) and the other from a cloudy day (cloudy). 

Gardens Point Walking dataset has provided frame-level correspondences. As for CMU and Mapillary datasets, we randomly select some frames and build frame-level correspondences as well. Details are summarized in Table~\ref{tab:t1}.

\subsubsection{\bf{Results}}
\begin{figure}[t]
\centering
\begin{tabular}{c}
\includegraphics[width=6.0cm]{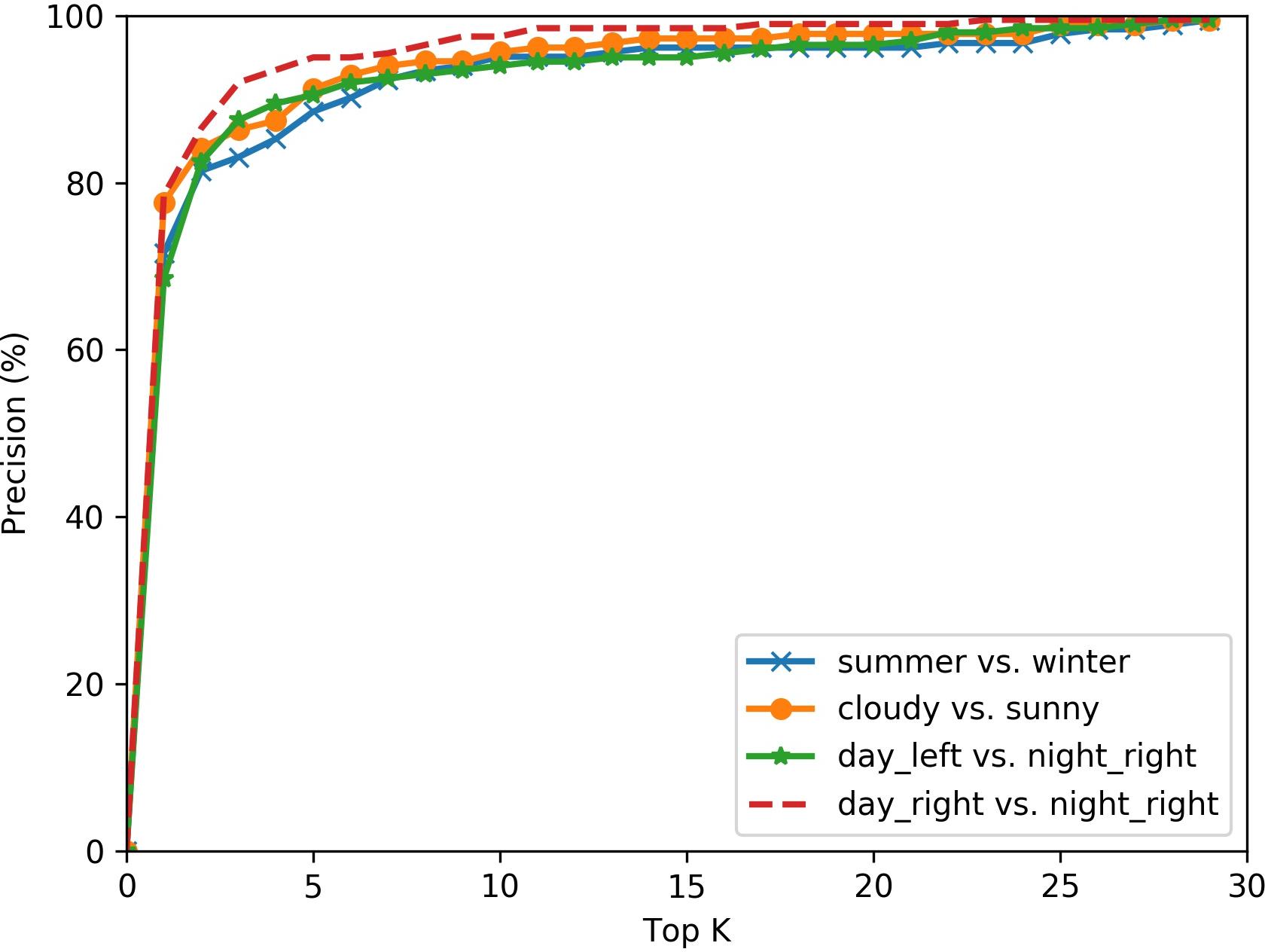}
\end{tabular}
\caption{The precision of the \textbf{Hostislic} descriptor with the increasing of top result numbers. Displaying the precision at 100\% recall in percentage format(\%).}
\label{fig:f4}
\vspace{-6.5mm}
\end{figure}

Fig.~\ref{fig:f4} shows the accuracy of the \textbf{Hostislic} descriptor with the increasing of top result numbers. When the top number is $30$, almost all reference map images can be retrieved. It is effective to eliminate unsimilar images for measurement with local features. 

\begin{figure}[t]
\centering
\begin{tabular}{cc}
\hspace{-5mm}
\includegraphics[width=4.3cm]{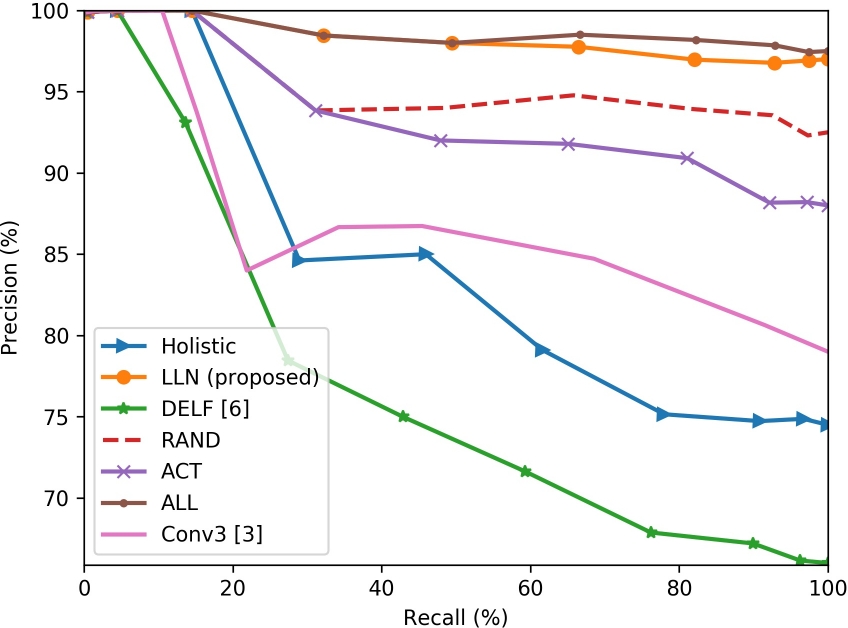}
\includegraphics[width=4.3cm]{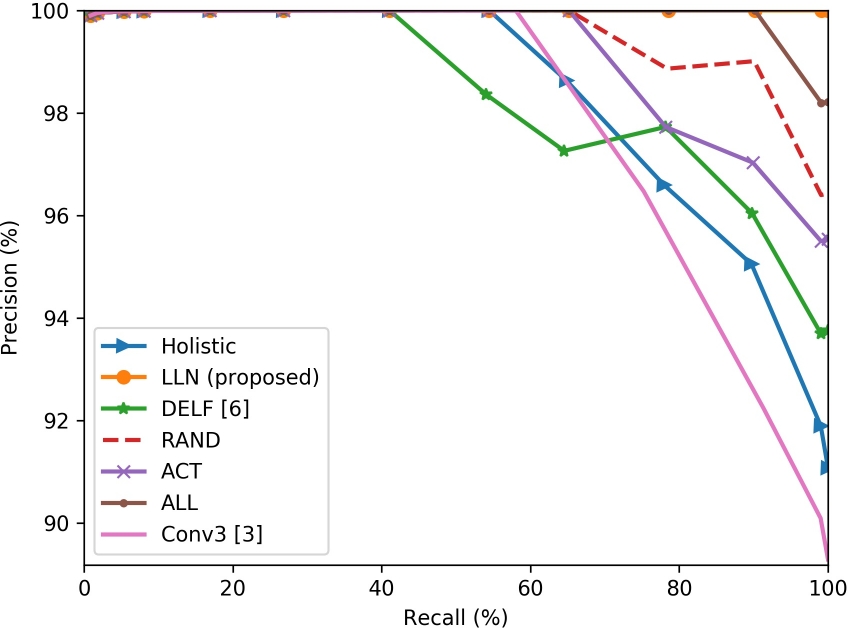}
\end{tabular}
\caption{The results of visual place recognition tested on day\_right vs. night\_right sequences (Gardens Point) and fall vs. winter sequences (CMU).}
\label{fig:f5}
\vspace{-2.5mm}
\end{figure}

Fig.~\ref{fig:f5} presents the Precision-Recall curves of all compared approaches. We can see that holistic features are outperformed by local features. Moreover, \textbf{LLN} achieves comparable even better performance than \textbf{ALL} with a small number of landmarks. With the same number of landmarks, \textbf{RAND} and \textbf{ACT} are outperformed by \textbf{LLN}.

As for \textbf{RAND}, we use the same random locations for both query and map images. This is a strong supervision when the testing datasets have moderate viewpoint changes since all landmarks in the query image can find their best corresponding landmarks in the map image. However, the randomly selected landmarks may not be discriminative, such as road surface or sky. These regions may sometimes lead to wrong recognition results, which makes its the performance worse than \textbf{LLN}. On the contrary, the proposed LLN can effectively localize discriminative image regions, which can not only make the similarity measurement much more efficient but also eliminate misleading regions. As for \textbf{ACT}, the base network is pre-trained on the task of object recognition, where the most active regions may not be suitable for representing a unique place.

\begin{figure}[t]
\centering
\begin{tabular}{c}
\includegraphics[width=8.0cm]{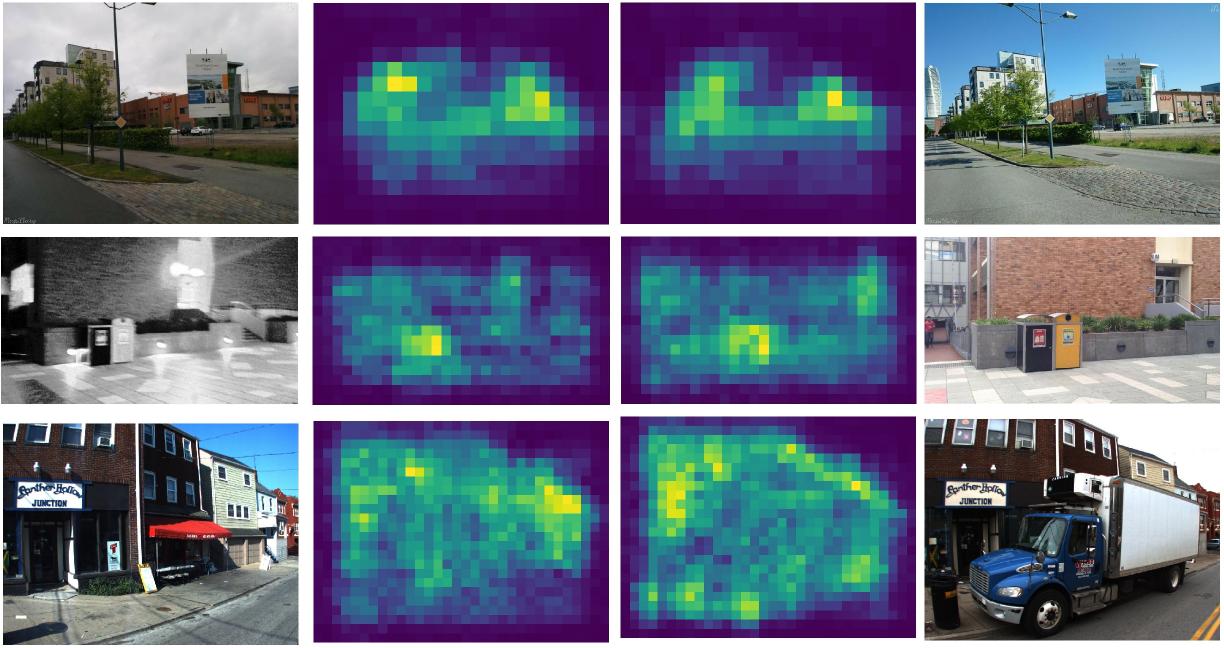}
\end{tabular}
\caption{Some matched image and their corresponding activation maps. Even with severe appearance changes, the proposed approach can find stable landmarks. For best visualization, images are fed into the network with their original resolution.}
\label{fig:f7}
\vspace{-5.5mm}
\end{figure}

As we can observe from Fig.~\ref{fig:f5}, \textbf{DELF} performs worse than \textbf{LLN} especially in Gardens Point Walking dataset. It may not generate landmark pairs in severe day-night changes. \textbf{DELF} jointly trains the attention network and local features. The training dataset used for image retrieval has no severe day-night appearance changes. In addition, \textbf{DELF} is trained with classification loss and has only single-scale attention filter. \textbf{LLN} utilizes metric learning to identify unique places and employs multi-scale filters to overcome partial occlusions. Fig.~\ref{fig:f7} shows some matched images and their activation maps generated from the proposed approach. More thorough comparisons are presented in Table~\ref{tab:t2}. 

\begin{table*}[t]
\caption{Results of place recognition. Displaying the precision at 100\% recall in percentage format(\%).}
\vspace{-4mm}
\label{tab:t2}
\begin{center}
\begin{tabular}{|c|c|c|c|c|c|c|c|} 
\hline 
\multirow{2}{*}{\bf{Sequences}} & \multicolumn{7}{c|}{\bf{Methods}} \\
\cline{2-8}
 & \bf{Holistic} & \bf{LLN} & \bf{DELF~\cite{delf}} & \bf{RAND} & \bf{ACT} & \bf{ALL} & \bf{Conv3~\cite{conv3}} \\
\hline
day\_left vs. night\_right & 63.5 & \bf{90.0} & 52.0 & 77.5 & 71.0 & 88.0 & 48.5 \\
\hline
day\_right vs. night\_right & 74.5 & 97.0 & 66.0 & 92.5 & 88.0 & \bf{97.5} & 79.0 \\
\hline
summer vs. fall & 89.3 & 96.4 & 95.5 & 96.4 & 95.5 & \bf{99.1} & 92.0  \\
\hline
summer vs. winter & 75.0 & 92.0 & 84.8 & 89.3 & 90.2 & \bf{94.6} & 82.0\\
\hline
fall vs. winter & 91.1 & \bf{100} & 93.8 & 96.4 & 95.5 & 98.2 & 89.3\\
\hline
cloudy vs. sunny & 77.6 & 95.1 & \bf{96.2} & 87.4 & 78.1 & 94.5 & 88.5\\
\hline
\bf{Average} & 78.5 & 95.1 & 81.4 & 89.9 & 82.1 & 95.3 & 79.9\\
\hline
\end{tabular}
\end{center}
\end{table*}

\subsection{Relocalization}
To further evaluate the robustness of the proposed approach. We integrate the whole process into ORB-SLAM to test the success rate of relocalization with severe environmental changes.

First, we utilize ORB-SLAM to build a map and eliminate the accumulated drift error based on GPS. All keyframes are treated as reference images for visual place recognition. Then, we turn on the localization mode and try to relocalization every test image in the map. There are two main components to relocalize the robot. One is searching candidate keyframes (visual place recognition), the other is keypoint feature matching (metric localization). We demonstrate the proposed approach can improve the performance of both steps. 

\subsubsection{\bf{Comparison and Evaluation Protocol}}
ORB-SLAM employs DBoW~\cite{dbow} to search candidate keyframes for place recognition (\textbf{pDBoW}) and also uses DBoW to constrain matched keypoints belonging to the same visual word in the vocabulary (\textbf{mDBoW}). 

We first replace the keyframe searching process with the proposed landmark localization approach (\textbf{pLLN}). Then, based on the mutually matched regions obtained in the similarity measurement, we further restrain keypoints to match in these regions (\textbf{mLLN}), as Xin et al.~\cite{xz}. The keypoints in a region of the query image find their matches only in the corresponding region of the recognized map image if the corresponding region exists. We resize the region to the original size before keypoint matching operation, since the images are resized to feed into the proposed network. 

Following the same strategy in ORB-SLAM, a successful relocalization is recorded if the 6-DOF pose is estimated after PnP, searching by reprojection and pose optimization process. $75$ landmarks are extracted for each query and map image. The top $5$ recognized keyframes are treated as candidates.

\subsubsection{\bf{Datasets}}
\begin{figure}[t]
\centering
\begin{tabular}{c}
\includegraphics[width=8.0cm]{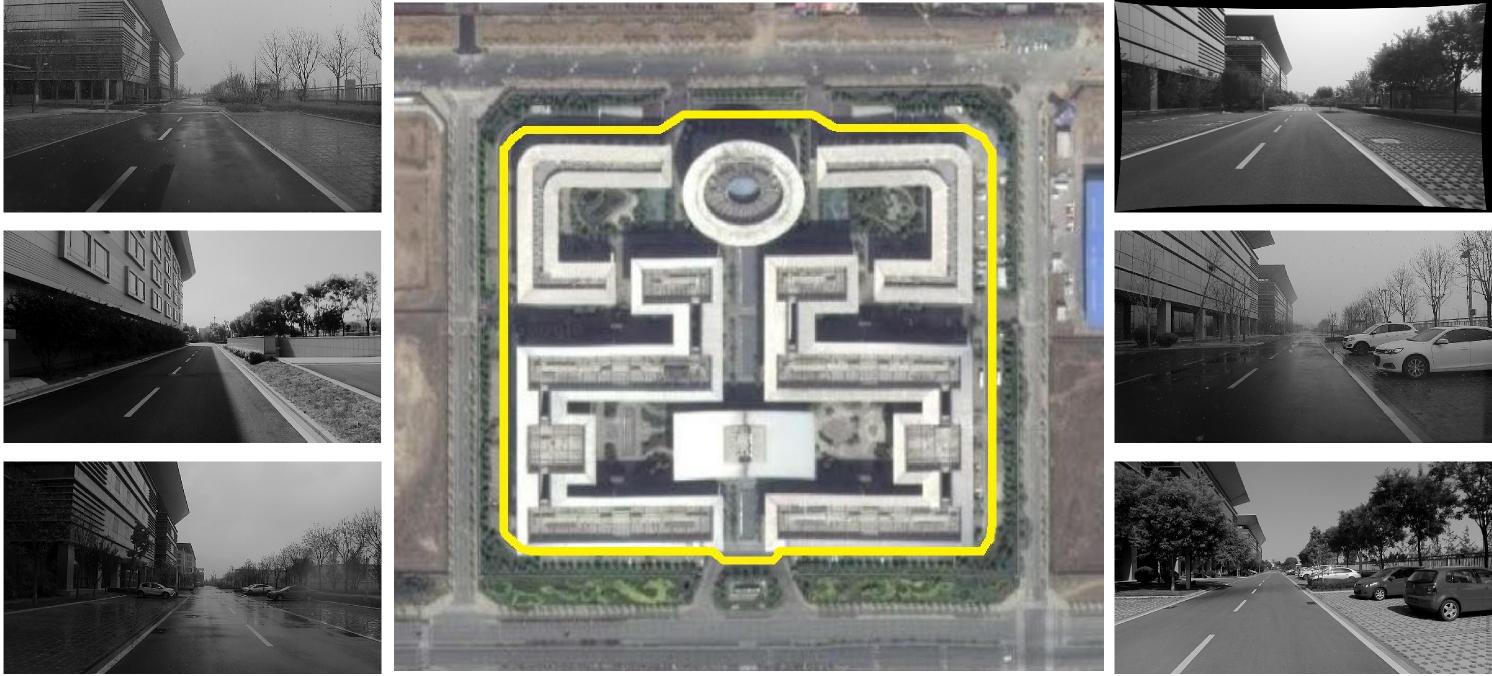}
\end{tabular}
\caption{Sample images of the dataset used for relocalization.}
\label{fig:f8}
\end{figure}

In this experiment, images captured under various environmental conditions are collected by a self-driving vehicle in a campus area. There are total 4 traverses. We use one traverse for mapping. The other traverses are used for relocalization. All traverses follow similar trajectories and have GPS information as the ground truth. Fig.~\ref{fig:f8} shows the route on a Google street-view map and some sample images. The characteristics of each traverse are summarized in Table~\ref{tab:t3}. The environmental changes include illumination changes, weather and season changes and dynamic objects such as pedestrians and vehicles. The resolution of the image is $1280\times 720$. All images are resized to $512 \times 288$ as the input to the proposed network.

\begin{table}[t]
\caption{The characteristics of relocalization traverses.}
\vspace{-4mm}
\label{tab:t3}
\begin{center}
\begin{tabular}{|ccccc|}
\hline 
\bf{Traverses} & \bf{Weather}& \bf{Time of Day}& \bf{Date}& \bf{Lanes}\\
\hline 
Map & Sunny & 9:30 & 2018/7/23 & Outside\\
\hline
Reloc1 & Sunny & 15:30 & 2018/5/18 & Outside\\
\hline
Reloc2 & Rain & 17:20 & 2018/4/13 & Inside\\
\hline
Reloc3 & Snow & 10:45 & 2018/3/17 & Outside\\
\hline
\end{tabular}
\end{center}
\vspace{-3mm}
\end{table}

\subsubsection{\bf{Results}}
\begin{figure}[t]
\centering
\begin{tabular}{c}
\includegraphics[width=8.0cm]{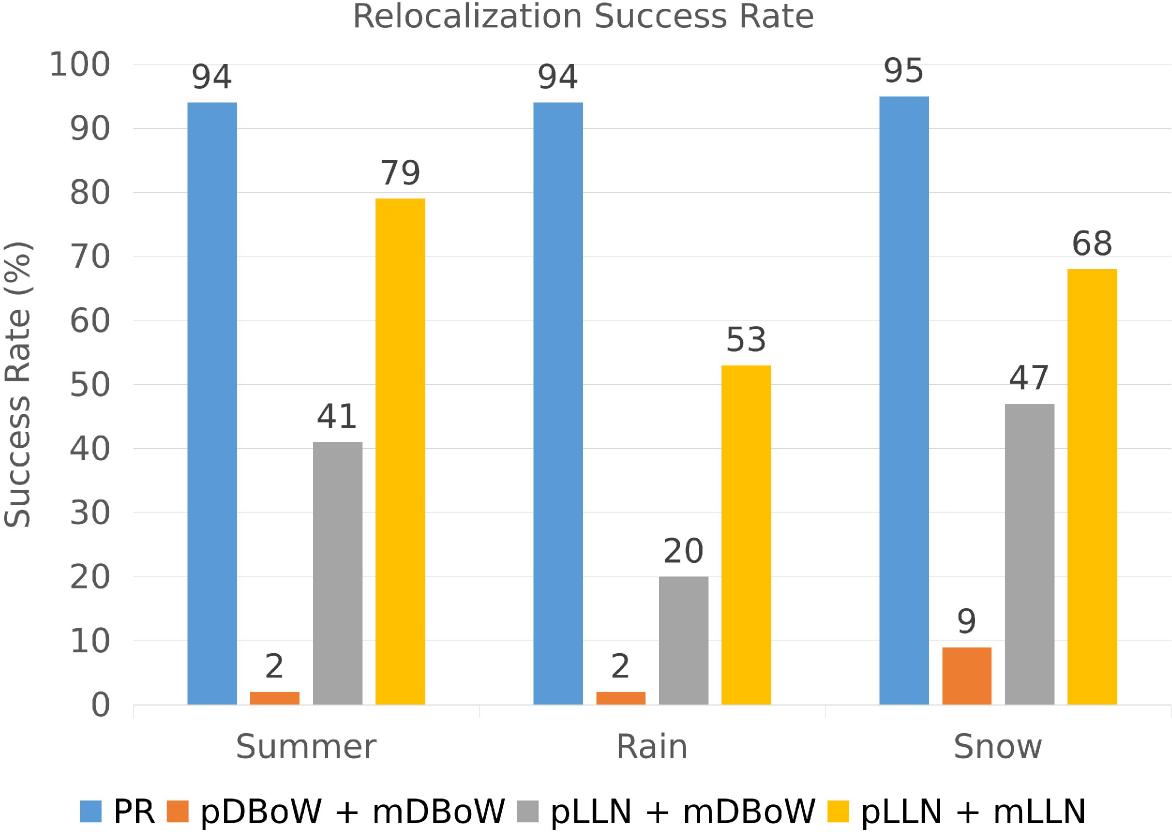}
\end{tabular}
\caption{The success rate of relocalization of test traverses.}
\label{fig:f9}
\end{figure}

Fig.~\ref{fig:f9} shows the success rate of all test traverses. The result of visual place recognition (\textbf{PR}) is performed by the proposed approach, which shows the upper bound of relocalization. The original ORB-SLAM (\textbf{pDBoW+mDBoW}) cannot handle severe appearance changes. Combining with LLN, more accurate candidates lead to a significant improvement in the results. \textbf{pLLN+mLLN} further improves the success rate, which is about two times of the \textbf{pLLN+mDBoW} results. LLN can effectively localize discriminative image regions. These regions can not only restrain the searching space of the matching process but also are much suitable for generating keypoint matches since discriminative regions generally contain rich texture information. The results in Fig.~\ref{fig:f9} demonstrate that the proposed approach does greatly boost the performance of relocalization in both visual place recognition and pose estimation.

However, the overall success rate is still not satisfactory. Even though LLN can achieve superior performance in image-based localization. The 6-DOF pose cannot be estimated robustly even with image region constraints. One reason is that ORB features are not robust against perceptual changes.

\subsection{Elapsed time analyzing}
We analyze the elapsed time of the proposed approach in Table~\ref{tab:t4}. The time of landmark extraction is computed per image. The speed of \textbf{Holistic}, \textbf{LLN} and \textbf{ALL} matching are computed per image pair. It is observed from Table~\ref{tab:t4} that \textbf{LLN} is much faster than \textbf{ALL} and barely losses any accuracy, which demonstrates the effectiveness and the efficiency of the proposed method.

\begin{table}[t]
\caption{Elapsed time of the proposed approach.}
\vspace{-4mm}
\label{tab:t4}
\begin{center}
\begin{tabular}{|cc|}
\hline 
\bf{Stage} & \bf{Elapsed time (No. of landmarks)}\\
\hline
Landmark extraction & 16 ms\\
\hline
{\bf{Holistic}} matching & 0.009 ms\\
\hline
{\bf{LLN}} matching & 6 (50) / 13 (75) ms\\
\hline
{\bf{ALL}} matching & 27 (108) / 46 (144) ms\\
\hline
\end{tabular}
\end{center}
\vspace{-5mm}
\end{table}

\section{CONCLUSIONS}
In this paper, we address the problem of visual place recognition in changing environments. A convolutional neural network (LLN) is designed to localize discriminative visual landmarks for representing images. The network is trained end-to-end only with image-level annotations. Detailed experiments demonstrate that the proposed approach achieves superior performance against state-of-the-art methods. We use an image retrieval dataset to learn relatively generic visual landmarks such as buildings and vegetables. The proposed method can also be particularly trained to find out useful landmarks for specialized environments.

\section*{ACKNOWLEDGMENT}
This work is supported by NSFC\#61503381 and the National Key R\&D Program of China (grant 2017YFB1300202).

\clearpage




%
%
%



\end{document}